\documentclass[a4paper,fleqn]{cas-dc}

\usepackage[numbers]{natbib}

\def\tsc#1{\csdef{#1}{\textsc{\lowercase{#1}}\xspace}}
\tsc{WGM}
\tsc{QE}

\usepackage{graphicx}
\usepackage{cite}
\usepackage{picinpar}
\usepackage{amsmath,amsfonts}
\usepackage{url}
\usepackage[utf8]{inputenc}
\usepackage{colortbl}
\usepackage{soul}
\usepackage{multirow}
\usepackage{pifont}
\usepackage{color}
\usepackage{alltt}
\usepackage{enumerate}
\usepackage{siunitx}
 
\usepackage{epstopdf}
\usepackage{pbox}
 
\usepackage{algorithm}
\usepackage{algorithmic}
\usepackage{booktabs}
\usepackage{tablefootnote}
\usepackage{hyperref}
 
\begin{document}
\let\WriteBookmarks\relax
\def\floatpagepagefraction{1}
\def\textpagefraction{.001}

\shorttitle{ImiPath}    


\title [mode = title]{Learning Spatiotemporal Decision Priors for Efficient Path Planning under Partial Observability}  


 

\author[1]{Yi Liu}[orcid=0000-0003-2221-2998]

\ead{liuyi_@fudan.edu.cn}

\author[1]{Hongda Zhang}
\author[1]{Leyao Zou}
\author[1]{Chunlei Meng}
\author[1]{Ziqing Zhou}

\author[1]{Yuning Chen}

\author[2]{Zhuo Zou}
\author[3]{Lida Xu}
\author[1]{Zhongxue Gan}

\author[1]{Chun Ouyang}
\cormark[1]
\ead{oy_c@fudan.edu.cn}

\cortext[cor1]{Corresponding author}

\affiliation[1]{organization={College of Intelligent Robotics and Advanced Manufacturing, Fudan University},
            city={Shanghai},
            postcode={200433}, 
            country={China}}

\affiliation[2]{organization={School of Information Science and Technology, Fudan University},
            city={Shanghai},
            postcode={200433}, 
            country={China}}

\affiliation[3]{organization={Department of Information Technology, Old Dominion University},
            city={Norfolk},
            postcode={23529}, 
            state={VA},
            country={USA}}

\begin{abstract}
Path planning under partial observability remains challenging because an agent must make long-horizon navigation decisions from only locally bounded observations. 
Nevertheless, historical trajectories contain reusable experience-guided directional preferences. 
Classical planners, however, typically solve each instance from scratch and lack an explicit mechanism to exploit such transferable decision knowledge, often leading to redundant node expansions and locally myopic search behaviors.
Motivated by this limitation, this paper proposes ImiPath, a prior-guided learning framework that distills reusable spatiotemporal decision priors from demonstration trajectories and uses them as experience-informed directional guidance to bias planners toward reliable and promising search directions under partial observability.
Specifically, ImiPath first constructs a local spatiotemporal observation representation, which encodes the spatial information of the local environment and the temporal information of historical trajectories. 
The SpatioTemporal-Attention Policy Network (STAPNet) then transforms this representation into dicision priors.
These priors are further incorporated into heterogeneous planners as directional guidance, biasing the search toward locally promising regions.
Extensive experiments demonstrate that ImiPath achieves competitive path quality and improves search efficiency by reducing redundant node expansions under local observability.
Additional physical experiments on a magnetic microrobot platform further validate the adaptability and practical deployment potential of the proposed framework. 

\end{abstract}








\maketitle

\section{Introduction}

Robot path planning~\citep{ugwoke2025simulation} aims to generate collision-free and feasible trajectories from a start position to a target while satisfying environmental and platform constraints. It serves as a fundamental component in autonomous robotic systems~\citep{reda2024path}, bridging perception, decision making, and motion execution. Path-planning techniques have been widely applied to mobile robots~\citep{chen2026integrated}, aerial vehicles~\citep{sathya2026realistic}, warehouse logistics~\citep{lai2026research}, search-and-rescue systems~\citep{zhu2025enhancing}, service robots, and microrobotic navigation~\citep{zou2025modified}. 

Classical path-planning methods can be broadly divided into deterministic and stochastic paradigms~\citep{liu2023path}. Deterministic planners, such as Dijkstra~\citep{ahmad2025enhancing} and A*~\citep{xu2024research}, provide structured search procedures and desirable theoretical properties under suitable assumptions. Stochastic planners, including sampling-based~\citep{ge2025rrt} and swarm-intelligence methods~\citep{cui2024multi, ou2026gpu}, provide stronger exploration capability and flexibility in complex environments, but they usually require extensive sampling, careful hyperparameter tuning, and planner-specific transition rules~\citep{cui2024multi}. 
However, their practical efficiency often depends on carefully designed heuristics, and their performance can degrade when global information is unavailable. 
Recent learning-based path-planning methods attempt to reduce the dependence on handcrafted heuristics by learning guidance signals from data~\citep{yonetani2021path}. Neural search methods and neural heuristic models~\citep{kirilenko2023transpath} have shown that learned cost maps, correction factors, or path probability maps can improve search efficiency in grid-based planning. Learning has also been introduced into stochastic planners~\citep{ren2026aco} to predict pheromone-related guidance or sampling distributions. 
Nevertheless, most existing methods are designed for globally observable settings or tightly coupled to specific planning algorithms, making their learned guidance signals difficult to reuse across heterogeneous planning paradigms~\citep{liu2023learning}.

Despite substantial progress in path planning, efficient and reliable planning remains difficult for real robotic systems under partial observability~\citep{ren2026aco}, where onboard sensors provide only bounded local observations and the obstacle layout beyond the current view remains unknown. In this setting, global heuristics or sampling strategies cannot be directly evaluated, forcing planners to rely on short-range goal cues and thereby causing redundant expansions, myopic decisions, future detours, or dead ends~\citep{zhou2024optimized}.
Therefore, a key challenge is how to learn reusable local decision priors that is independent of global map scale, compatible with different planning mechanisms, and effective for robust decision making under partial observability.
It is inspired by human behavior in finding their way: even without access to a complete map, humans can often exploit prior navigation experience to prefer more promising directions based on local spatial layouts, approximate goal directions, and recent movement history.
This observation motivates the learning of transferable local decision priors for planning under partial observability.

To address this challenge, this paper proposes ImiPath, an imitation learning-based framework that distills reusable spatiotemporal decision priors from expert demonstrations for path planning under local observability.
Specifically, the spatiotemporal observation representation is constructed to encode both the spatial information of the local environment and the temporal information of historical trajectories. Based on this representation, the SpatioTemporal-Attention Policy Network (STAPNet) is developed to predict the
policy distribution, which serves as the spatiotemporal dicision prior. 
Rather than treating the learned policy as a standalone planner, 
exploits the reusable experience-guided decision prior to guide the search process. 
This design mitigates the vulnerability of purely learned policies to prediction uncertainty and accumulated errors under partial observability, while preserving the structured search mechanism and feasibility reasoning of classical planners

The main contributions are summarized as follows:
\begin{itemize}
    \item 
    ImiPath is proposed as a prior-guided learning framework that distills reusable spatiotemporal decision priors from expert demonstrations for path planning under partial observability.

\item STAPNet learns experience-guided spatiotemporal decision priors from large-scale demonstration trajectories and formulates them as directional guidance for heterogeneous search algorithms, enabling different planners to reuse expert-derived decision knowledge and reduce redundant exploration.

    \item Extensive evaluations across multiple map scales, dynamic scenarios, and a magnetic microrobot platform demonstrate that ImiPath achieves competitive path quality, reduces redundant node expansions, enables millisecond-level prior inference, and supports closed-loop navigation from sequential local observations under partial observability.
\end{itemize}

\section{Related Work}
\label{sec:relatedwork}

\subsection{Classical deterministic and stochastic planners}

Classical path-planning methods are commonly categorized into deterministic and stochastic paradigms. 
Deterministic planners, such as Dijkstra and A*-based methods~\citep{rao2023path}, perform structured graph search and can provide desirable completeness or optimality properties under appropriate assumptions. 
Their practical performance, however, largely depends on the design of heuristic functions and cost models, particularly in complex environments~\citep{jeon2024poster}.
Considerable efforts have therefore been devoted to improving their search mechanisms~\citep{lin2023efficient}.
For example, Xu et al.~\citep{xu2024research} adopted adaptive cost functions and enhanced map representations for global mobile robot planning, while Huang et al.~\citep{xu2024research} proposed a self-adaptive neighborhood A* algorithm to reduce search redundancy in cluttered environments. 
Despite these advances, deterministic methods still face difficulties when extended to settings with limited sensing and partial observability~\citep{harabor2019regarding}.

Stochastic planners enhance exploration through probabilistic transitions or sampling mechanisms and often provide stronger flexibility in irregular or highly constrained environments. They can be broadly divided into sampling-based methods, such as PRM and RRT*, and graph- or transition-based metaheuristic planners, such as ACO~\citep{wu2023modified}. Since the target problem is partially observable grid navigation with discrete actions, this study focuses on ACO-based stochastic planners, whose transition rules are more amenable to prior fusion. Representative methods in this category include IHMACO~\citep{zhao2022ant}, which incorporates evolutionary experience-guided pheromone updates, and PFACO~\citep{liu2026pheromone}, which improves efficiency and path quality through targeted pheromone placement and iterative exploitation of high-quality paths. Nevertheless, such methods usually require careful parameter tuning and may suffer from limited deployment efficiency and weak transferability across tasks and environments.

Overall, although deterministic and stochastic planners employ different search mechanisms, they share a common limitation: their guidance signals are predominantly manually designed and planner-specific. 
As a result, they typically solve each planning instance independently rather than reusing prior decision knowledge. 
This limitation becomes particularly pronounced under partial observability, where reusable local decision priors may be more valuable than handcrafted global search rules.

 \begin{figure*}[t]
  \centering
  \includegraphics[width=0.97\linewidth]{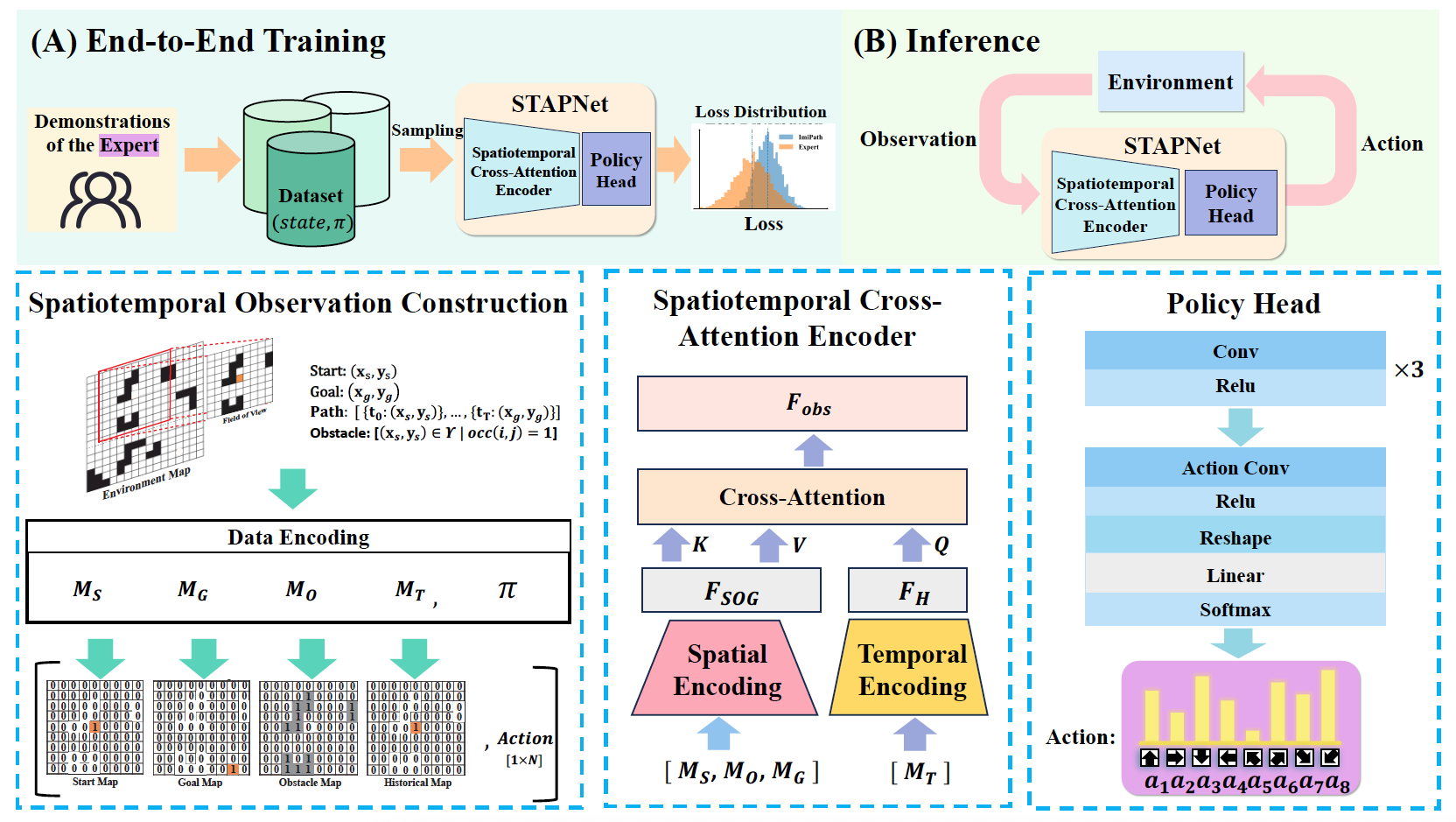}
  \caption{The Framework of ImiPath.
Top: The pipeline consists of end-to-end training (A) and inference (B). During training, expert demonstrations are encoded into spatiotemporal observations and used to learn decision priors via the STAPNet module. The STAPNet integrates a spatiotemporal cross-attention encoder and a policy head. At inference time, the learned policy operates under partial observability to guide planning.
Bottom: the components of observation construction, spatiotemporal cross-attention encoder, and policy head.
  }
  \label{fig:processImipath}
\end{figure*}

\subsection{Learning-Based Path Planning and Neural Heuristics}

In response to the limitations of classical planners, learning-based approaches have emerged as promising alternatives for path planning. These methods reduce dependence on handcrafted heuristics and leverage data-driven models to capture problem-dependent search patterns~\citep{liu2023path}. 
A major line of work focuses on learned guidance for classical search. Neural A* Search~\citep{yonetani2021path} incorporates differentiable neural modules into the search process, enabling end-to-end learning of search-favorable transition costs. TransPath~\citep{kirilenko2023transpath} further extends this idea by learning heuristic proxies, including a correction factor for weighted A* (WA*+CF) and a path probability map for focal search (FS+PPM), thereby improving search efficiency on fully observable grid maps.

Learning has also been incorporated into stochastic planning. 
In ACO-based methods~\citep{ren2026aco}, neural models have been used to predict pheromone-related guidance or search distributions and incorporate them into global pheromone-based exploration ~\citep{liu2023learning, ye2023deepaco}.
Such hybrid methods combine the exploration capability of stochastic planners with the pattern-recognition ability of neural networks. 
However, many of them remain tightly coupled to a specific planner and often rely on globally available environmental information. 
Consequently, the learned guidance may not transfer readily across heterogeneous planning paradigms or to scenarios where only local observations are available.

Another related direction is imitation learning for planning, where policies are learned from expert demonstrations to map observations directly to actions \citep{bhardwaj2017learning}. 
Imitation learning can capture expert-like decision patterns and enable efficient inference, but a standalone learned policy may suffer from accumulated errors or reduced robustness when deployed in unseen environments. 
This issue becomes more pronounced under partial observability, where short-term local decisions must remain consistent with long-horizon planning objectives. 
Therefore, rather than replacing classical planners with a purely learned policy, it is desirable to use the learned policy as reusable decision knowledge that complements classical search.

\section{Proposed Method}\label{sec:Imipath}

ImiPath is a prior-guided learning framework for path planning under partial observability.
Its main component, the STAPNet, learns historical decision distributions from expert demonstrations, which serve as the spatiotemporal dicision prior and guide the search process of downstream planners under partial observability.

As illustrated in Fig.~\ref{fig:processImipath}, ImiPath consists of a training phase and an inference phase. 
Expert trajectories are generated by applying advanced path-planning algorithms to large-scale planning tasks across diverse environments. 
During training, these trajectories are converted into state--policy pairs $(\textit{state}, \pi)$, providing experience data from which STAPNet learns spatiotemporal decision priors.
During inference, STAPNet operates on local observations in a closed loop and outputs decision priors that can be used for guiding other classical planners.

\begin{figure} 
  \centering
  \includegraphics[width=0.7\columnwidth]{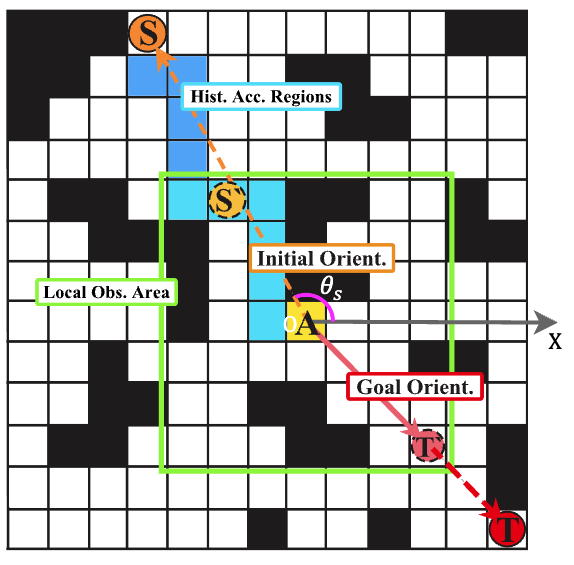}
  \caption{
  Illustration of the proposed local spatiotemporal observation. The green box denotes the local FoV centered at the current agent position \(A\). White and black cells represent free space and obstacles, respectively. The global start \(S\) and goal \(T\) are marked by orange and red circles. When they lie outside the current FoV, their relative directions are preserved by projecting them onto the FoV boundary as \(S'\) and \(T'\). Light blue cells indicate recently visited regions within the FoV, while deep blue cells denote earlier trajectory segments outside the current view. The polar coordinate system is centered at the agent position \(O\).
  }
  \label{fig:observation}
\end{figure}

\subsection{Spatiotemporal Observation Construction}
The path planning problem is formulated in a partially observable discrete grid world, where the agent can access only a local observation at each time step. In this work, the local FoV is defined as an \(11\times11\) grid centered at the agent. As illustrated in Fig.~\ref{fig:observation}, the proposed observation is constructed by extracting a local window around the current agent position and augmenting it with directional cues of the global start and goal, as well as recent trajectory history. To encode the relative directions of the start and goal in a spatially consistent manner, a polar coordinate system centered at the agent is introduced within the constrained FoV.
When the start or goal lies outside the current FoV, its relative direction is preserved by projecting it onto the FoV boundary as \(S'\) or \(T'\). In this way, the resulting observation jointly encodes local obstacle layout, projected start and goal cues, and temporally decayed motion history, thereby providing structured spatiotemporal cues for planning under limited observability.

\subsubsection{Global-to-Local Coordinate Mapping}

Let the current agent position, the start position, and the goal position be denoted by \(P_a=(x_a,y_a)\), \(P_s=(x_s,y_s)\), and \(P_g=(x_g,y_g)\), respectively. Their relative displacements with respect to the agent are defined as
\begin{equation}
\Delta x_s = x_s - x_a, \quad \Delta y_s = y_s - y_a,
\end{equation}

\begin{equation}
\Delta x_g = x_g - x_a, \quad \Delta y_g = y_g - y_a.
\end{equation}

These displacements are then expressed in polar form as
\begin{align}
r_s &= \sqrt{(\Delta x_s)^2 + (\Delta y_s)^2}, &
\theta_s &= \arctan2(\Delta y_s, \Delta x_s), \\
r_g &= \sqrt{(\Delta x_g)^2 + (\Delta y_g)^2}, &
\theta_g &= \arctan2(\Delta y_g, \Delta x_g),
\end{align}
where \(r_s\) and \(r_g\) denote the Euclidean distances from the current agent position to the start and goal, respectively, and \(\theta_s\) and \(\theta_g\) denote the corresponding directions.

To determine whether a node lies inside the FoV, the maximum observable radius is defined as
\begin{equation}
r_{\max}=\frac{W-1}{2},
\end{equation}
where \(W\) is the width of the local observation window.

When the start lies outside the observation window, i.e., \(r_s>r_{\max}\), it is projected onto the FoV boundary as
\begin{equation}
\begin{aligned}
P'_s &= (\tilde{x}_s,\tilde{y}_s) \\
     &= \left(
\frac{W-1}{2}+r_{\max}\cos\theta_s,\;
\frac{W-1}{2}-r_{\max}\sin\theta_s
\right),
\end{aligned}
\end{equation}
and the projected goal position is similarly defined as
\begin{equation}
\begin{aligned}
P'_g &= \left(\tilde{x}_g,\tilde{y}_g\right) \\
     &= \left( \frac{W-1}{2}+r_{\max}\cos\theta_g,\;
     \frac{W-1}{2}-r_{\max}\sin\theta_g \right).
\end{aligned}
\label{p_g}
\end{equation}
For observation encoding, the original position is retained when the start or goal lies within the current FoV; otherwise, its boundary projection is used to preserve the corresponding relative direction under limited observability.

\subsubsection{Spatiotemporal Observation Representation}
The local observation is encoded by four matrices: the start cue $M_S$, goal cue $M_G$, obstacle map $M_O$, and trajectory-history map $M_T$. 
For a local window of size $W \times W$, each matrix element $(i,j)$ is defined by the indicator function \(\mathbb{I}[\cdot]\), which returns \(1\) if the condition is satisfied and \(0\) otherwise:
\begin{align}
M_S(i,j) &= \mathbb{I}\big[(x_i,y_j)=(\tilde{x}_s,\tilde{y}_s)\big], \\[4pt]
M_G(i,j) &= \mathbb{I}\big[(x_i,y_j)=(\tilde{x}_g,\tilde{y}_g)\big], \\[4pt]
M_O(i,j) &= \mathbb{I}\big[(x_i,y_j)\in\mathcal{O}\big], \\[4pt]
M_T(i,j) &= \sum_{k=0}^{T}\gamma^{T-k}\mathbb{I}\big[(x_i,y_j)=(x_k,y_k)\big],
\end{align}
where $\mathcal{O}$ denotes the set of obstacle cells, $(x_k,y_k)$ is the agent position at time step $k$, and $\gamma \in (0,1]$ is a temporal decay factor that assigns larger weights to more recent trajectory states.
Accordingly, the local observation is represented as
\begin{equation}
\textit{state}=[M_S,M_O,M_G,M_T],
\end{equation}
which jointly encodes local geometry, directional information, and motion history within the fixed FoV.

\subsection{Action Space}

The agent operates in an eight-connected discrete action space and can move to any of its eight neighboring cells at each time step. The action set is defined as
\begin{align*}
\mathcal{A}=\{(x,y)\mid x,y\in\{-1,0,1\},\ (x,y)\neq(0,0)\}.
\end{align*}

For the dynamic-environment experiments, the action space is extended with an additional~\emph{stop} action to allow temporary waiting when a moving obstacle creates an imminent collision risk.

\subsection{SpatioTemporal-Attention Policy Network (STAPNet)}
\label{subsec:STAPNet}

As shown in Fig.~\ref{fig:processImipath}, this network consists of a spatiotemporal cross-attention encoder and a policy head.

\subsubsection{Spatiotemporal Cross-Attention Encoder}

To learn spatiotemporal decision priors under partial observability, the encoder jointly models task-relevant spatial context and historical trajectories. 
The spatial components $M_S$, $M_G$, and $M_O$ are encoded into a spatial feature representation $F_{SOG}$, while the trajectory-history map $M_T$ is encoded into a temporal feature representation $F_H$. 
Cross-attention is then applied by using $F_H$ as the query and $F_{SOG}$ as both the key and value, yielding the fused observation representation $F_{\mathrm{obs}}$. 
This design uses historical trajectories to encode previously visited regions, allowing the model to retrieve decision-relevant spatial cues, discourage repeated exploration, and generate decision priors conditioned on both local structure and recent traversal history.

\subsubsection{Policy Head}
 
The fused representation $F_{\mathrm{obs}}$ is further processed by the policy head to produce the policy distribution. 
For each admissible action $a_i\in\mathcal{A}$, its probability is computed as
\begin{equation}
\varphi_{\theta}(a_i \mid \textit{state})
=
\frac{\exp\left(f_{\theta}(F_{\mathrm{obs}})_i\right)}
{\sum_{j=1}^{N}\exp\left(f_{\theta}(F_{\mathrm{obs}})_j\right)},
\end{equation}
where $f_{\theta}(\cdot)$ denotes the output logits of the policy head and $N=|\mathcal{A}|$ is the number of admissible actions.

\subsection{Training and Inference}

The training procedure of STAPNet is summarized in Algorithm~\ref{alg:STAPNet_train}. Given the expert demonstration dataset
\begin{equation}
\mathcal{D}=\{(M_S^i,M_O^i,M_G^i,M_T^i,\pi_i)\mid i\in\mathcal{T}\},
\end{equation}
Here, $\pi_i$ denotes the expert target distribution over the action set at sample $i$; when each demonstration state is associated with a single expert action, $\pi_i$ is represented as a one-hot distribution.

The network parameters $\theta$ are optimized using AdamW with learning rate $\eta$ and weight decay $\lambda$. 
For a mini-batch of size $B$, the training objective is defined as
\begin{equation}
\label{eq:loss}
\mathcal{L}(\theta)=
-\frac{1}{B}\sum_{i=1}^{B}\sum_{a\in\mathcal{A}}
\pi_i(a)\log \varphi_\theta(a\mid state_i)
+\lambda\lVert\theta\rVert_2^2.
\end{equation}
where the first term is the cross-entropy loss for imitation learning and the second term is the $\ell_2$ regularization induced by weight decay.
 
During inference, STAPNet receives the current local observation
$(M_S^t,M_O^t,M_G^t,M_T^t)$ and outputs the policy distribution $\boldsymbol{\varphi}^t$. 
When STAPNet is evaluated as a standalone policy baseline, the action with the highest probability is selected as
\begin{equation}
a^t=\arg\max_{a\in\mathcal{A}}\boldsymbol{\varphi}^t[a].
\end{equation}
By repeating this procedure over time, STAPNet performs closed-loop local decision making. 
Within the complete ImiPath framework, however, $\boldsymbol{\varphi}^t$ is not used to directly determine the final path; instead, it serves as a learned spatiotemporal decision prior that is integrated into downstream planners, as described next.

\begin{algorithm}[!t]
\caption{STAPNet Training}
\label{alg:STAPNet_train}
\begin{algorithmic}[1]
\REQUIRE Dataset $\mathcal{D}=\{(M_S,M_O,M_G,M_T,\pi)\}$; action set $\mathcal{A}$; learning rate $\eta$; weight decay $\lambda$; batch size $B$; epochs $T$.
\STATE Initialize STAPNet parameters $\theta$.
\STATE Initialize the AdamW optimizer with learning rate $\eta$ and weight decay $\lambda$.
\STATE \textbf{Define} \textsc{Forward}$(M_S,M_O,M_G,M_T)$:
\STATE \hspace{0.6cm}$F_{SOG}\leftarrow \textsc{SpatialEncoding}(M_S,M_O,M_G)$
\STATE \hspace{0.6cm}$F_H\leftarrow \textsc{TemporalEncoding}(M_T)$
\STATE \hspace{0.6cm}$F_{\text{obs}}\leftarrow \textsc{CrossAttn}(Q{=}F_H,\;K{=}F_{SOG},\;V{=}F_{SOG})$
\STATE \hspace{0.6cm}$\boldsymbol{\varphi}\leftarrow \textsc{PolicyHead}(F_{\text{obs}})\in\mathbb{R}^{B\times|\mathcal{A}|}$
\STATE \hspace{0.6cm}\textbf{return} $\boldsymbol{\varphi}$
\FOR{$t=1$ to $T$}
    \FOR{minibatch $\{(M_S,M_O,M_G,M_T,\pi)\}_{b=1}^B \sim \mathcal{D}$}
        \STATE $\boldsymbol{\varphi} \leftarrow \textsc{Forward}\{(M_S,M_O,M_G,M_T)\}_{b=1}^B$
        \STATE $\mathcal{L} \gets \mathrm{CrossEntropy}(\boldsymbol{\pi},\boldsymbol{\varphi})$
        \STATE $\mathrm{zero\_grad}()$; \quad $\nabla_\theta \mathcal{L} \gets \mathrm{backward}(\mathcal{L})$
        \STATE $\mathrm{step}()$
    \ENDFOR
\ENDFOR
\end{algorithmic}
\end{algorithm}

\begin{algorithm}[!t]
\caption{STAPNet Inference}
\label{alg:STAPNet_inference}
\begin{algorithmic}[1]
\STATE \textbf{Input:} a single state \((M_S^t,M_O^t,M_G^t,M_T^t)\) at time \(t\)
\STATE $\boldsymbol{\varphi}^t \leftarrow STAPNet(M_S^t,M_O^t,M_G^t,M_T^t)$
\STATE $a^t \leftarrow \arg\max_{a\in\mathcal{A}}\boldsymbol{\varphi}^t[a]$
\STATE \textbf{return} $a^t$
\end{algorithmic}
\end{algorithm}

\subsection{Prior-Guided Planning Strategy}

The learned spatiotemporal decision prior is incorporated into downstream planners in a planner-compatible manner.  
The experience-guided spatiotemporal decision priors guides the native search mechanism of each planner by providing reusable local decision knowledge under partial observability.

\subsubsection{Deterministic Planning}
For deterministic planners such as A*, a globally informed heuristic cannot be directly evaluated under strict local FoV constraints, because the final goal may lie outside the current observation window and the unseen obstacle layout is unavailable. 
The local guidance term is defined as
\begin{equation}
h_{\mathrm{local}}(n)=d(n, g_{\mathrm{local}}),
\label{eq:local_heuristic}
\end{equation}
where $g_{\mathrm{local}}$ denotes the actual goal cell when the goal is visible; otherwise, it denotes the projected goal position $P'_g$ defined in Eq.~(\ref{p_g}). 
The function $d(\cdot,\cdot)$ is implemented as the Manhattan distance on the discrete grid and is used only as a simple local directional cue rather than a globally admissible heuristic.

For deterministic search, the prior provides as a local preference in node evaluation, encouraging expansions that are more consistent with expert-like local behavior. 
The resulting deterministic evaluation rule is formulated as
\begin{equation}
f'(n)=g(n)+\rho h_{\mathrm{local}}(n)-\mu \log \bigl(\varphi(a_n \mid state_n)+\epsilon\bigr),
\label{eq:deterministic_eval}
\end{equation}
where $g(n)$ denotes the accumulated path cost, 
$\varphi_{\theta}(a_n\mid \textit{state}_n)$ is the STAPNet-predicted probability of the candidate action leading to node $n$, 
$\rho$ and $\mu$ are weighting factors, and $\epsilon>0$ is a small constant for numerical stability. 
The logarithmic prior term assigns lower evaluation costs to actions with higher predicted preference, thereby introducing expert-informed local guidance while preserving explicit goal-directed search within the observable region.

\subsubsection{ACO-based Stochastic Planning}
For stochastic planning, the prior modulates the transition probabilities of candidate moves while preserving the original pheromone and heuristic components. 
This formulation allows the learned decision prior to be naturally incorporated as a local transition preference within the ACO search process.
Specifically, the STAPNet prediction is fused into the transition rule at each iteration so that pheromone-based sampling is continuously biased toward locally promising actions. 
The transition probability for ant $k$ moving from node $i$ to node $j$ is defined as
\begin{equation}\label{eq:pacotransfer}
p_{ij}^k =
\begin{cases}
\dfrac{\varphi_{ij}\tau_{ij}^{\alpha}\eta_{ij}^{\beta}}
{\sum_{l\in allowed_k}\varphi_{il}\tau_{il}^{\alpha}\eta_{il}^{\beta}}, & j\in allowed_k, \\[10pt]
0, & \text{otherwise},
\end{cases}
\end{equation}
where $\tau_{ij}$ denotes the pheromone value on edge $(i,j)$, 
$\eta_{ij}=1/d_{ij}$ is the distance-based heuristic factor, 
$\varphi_{ij}$ is the STAPNet-predicted probability associated with the local action from node $i$ to node $j$, and 
$\alpha$ and $\beta$ control the relative influence of pheromone and heuristic information, respectively. 
The next move is then sampled according to this transition distribution. 
In this way, the prior is iteratively injected into stochastic search without replacing the original pheromone accumulation mechanism.

By extracting priors from agent-centered local observations and task-relevant spatiotemporal cues rather than algorithm-specific search structures, this design enables general spatiotemporal decision priors to guide heterogeneous planning paradigms.

\section{Experiments}\label{sec:experiments}

\subsection{Simulation Setup}
\label{sec:simulation_setup}

All experiments were implemented in PyTorch under Python 3.7 and conducted on a Linux workstation running Ubuntu 20.04.6 LTS, equipped with an Intel Core i7-8700 CPU at 3.20 GHz and an NVIDIA GeForce RTX 3080 SUPER GPU.

A benchmark dataset was constructed by integrating maps from prior studies~\citep{bhardwaj2017learning, yonetani2021path, liu2023learning}. 
The dataset was split into training, validation, and test sets at the map-instance level. In this work, PFACO~\citep{liu2026pheromone} and $A^*_{\mathrm{Global}}$ were selected as expert planners to generate expert trajectories, as they provide high-quality planning results and serve as representative advanced methods in stochastic and deterministic planning, respectively.
STAPNet was trained exclusively on $11\times11$ local FoV observations from the training split and is directly evaluated, without fine-tuning, retraining, or parameter re-selection, on held-out test maps spanning different global sizes, including Small scale maps ($15 \times15$), Medium-scale map ($20 \times20$), Medium-large-scale map ($25 \times25$), Large-scale map ($30 \times30$). 
 
Both stochastic and deterministic planners were considered in the evaluation. 
For stochastic comparisons, representative ACO-based methods were selected, including AS~\citep{blum2005ant}, Elite AS (EAS)~\citep{wu2024adapted}, MMAS and MMAS2020~\citep{skinderowicz2020implementing}, IHMACO \citep{zhao2022ant}, and PFACO~\citep{liu2026pheromone}. 
Each stochastic planner is reported under two standardized parameter configurations. 
The subscript ``light'' denotes a lightweight setting with a population size of 15 and 10 search iterations, whereas the notation without a subscript denotes the full setting with a population size of 30 and 20 search iterations. 
The same convention is used for all stochastic baselines.

For deterministic comparisons, $A^*_{\mathrm{Local}}$, $A^*_{\mathrm{Global}}$, FS+PPM, and WA*+CF \citep{kirilenko2023transpath} were included. 
Among them, $A^*_{\mathrm{Local}}$ operates under the same local-FoV constraint as $\mathrm{ImiPath}_{\mathrm{A}^{*}_{\mathrm{Local}}}$ and therefore serves as the primary deterministic baseline for fair comparison under partial observability. 
By contrast, FS+PPM and WA*+CF assume globally available map information and are reported only on the FoV-scale benchmark as globally informed reference methods rather than strictly comparable baselines. 

Performance was evaluated using average path length (APL), average computation time per planning instance (Time), and success rate (SR). 
Here, SR denotes the percentage of successful goal-reaching trials among 100 test instances, APL is reported as mean $\pm$ standard deviation over successful trials, and Time is measured in seconds per planning instance. 
Lower APL and shorter Time indicate better path quality and computational efficiency, respectively, whereas higher SR indicates stronger planning robustness. 
For deterministic planners, the number of explored nodes is additionally reported to assess search efficiency.

To assess statistical significance in path quality, Wilcoxon signed-rank tests were conducted on the path-length distributions at $\alpha=0.05$, and the resulting outcomes are reported in the $\mathrm{APL}_{p}$ column. 
In the tables, ``+'' indicates that the baseline significantly outperforms ImiPath, ``$-$'' indicates that ImiPath significantly outperforms the baseline, ``$\approx$'' indicates no statistically significant difference, and ``$\times$'' indicates that the test was not conducted because the number of successful trials was insufficient for a reliable paired comparison.

\begin{table}[width=\linewidth,cols=5,pos=t]
\caption{Comparison of STAPNet, ImiPath, and representative stochastic path planners on small and medium map scales under local FoV constraints.}
\label{tab:aco_small_medium}

\begin{tabular*}{\tblwidth}{@{} LCCCC @{}}
\toprule
\textit{Method} & \textit{APL $\pm$ SD} & \textit{Time(s)} & \textit{SR(\%)} & \textit{$\mathrm{APL}_{\mathrm{p}}$} \\
\midrule

\multicolumn{5}{@{}l}{\textit{Map Scale: FoV Scale}} \\
\midrule
$\mathrm{AS}_{\mathrm{light}}$        & $5.863 \pm 4.494$   & $2.506 \times 10^{-1}$   & $100$ & $-$ \\
$\mathrm{AS}$                         & $5.502 \pm 4.032$   & $1.012$                   & $100$ & $-$ \\
$\mathrm{EAS}_{\mathrm{light}}$       & $5.052 \pm 3.522$   & $2.493 \times 10^{-1}$    & $100$ & $-$ \\
$\mathrm{EAS}$                        & $4.879 \pm 3.263$   & $9.988 \times 10^{-1}$    & $100$ & $\approx$ \\
$\mathrm{MMAS}_{\mathrm{light}}$      & $5.068 \pm 3.518$   & $2.561 \times 10^{-1}$    & $100$ & $-$ \\
$\mathrm{MMAS}$                       & $4.920 \pm 3.363$   & $1.053$                   & $100$ & $-$ \\
$\mathrm{MMAS2020}_{\mathrm{light}}$  & $4.766 \pm 3.865$   & $6.005 \times 10^{-2}$    & $51$  & $\times$ \\
$\mathrm{MMAS2020}$                   & $3.875 \pm 3.435$   & $1.241 \times 10^{-1}$    & $42$  & $\times$ \\
$\mathrm{IHMACO}_{\mathrm{light}}$    & $7.926 \pm 5.579$   & $1.517 \times 10^{-1}$    & $93$  & $\times$ \\
$\mathrm{IHMACO}$                     & $7.740 \pm 5.767$   & $2.419$                   & $94$  & $\times$ \\
$\mathrm{PFACO}_{\mathrm{light}}$     & $4.956 \pm 3.435$   & $9.461 \times 10^{-2}$    & $100$ & $-$ \\
$\mathrm{PFACO}$                      & $4.775 \pm 3.106$   & $3.132 \times 10^{-1}$    & $100$ & $\approx$ \\
$\mathrm{STAPNet}$                    & $5.127 \pm 3.950$   & $3.361 \times 10^{-3}$    & $99$  & $\approx$ \\
$\mathrm{ImiPath}^{\mathrm{light}}_{\mathrm{PFACO}}$   & $4.795 \pm 3.141$   & $\mathbf{1.160 \times 10^{-1}}$    & $100$ & $\approx$ \\
$\mathrm{ImiPath}_{\mathrm{PFACO}}$                    & $\mathbf{4.766 \pm 1.580}$   & $3.344 \times 10^{-1}$    & $100$ &  \\
\midrule

\multicolumn{5}{@{}l}{\textit{Map Scale: } Small} \\
\midrule
$\mathrm{AS}_{\mathrm{light}}$        & $12.167 \pm 6.383$  & $8.093 \times 10^{-1}$    & $100$ & $-$ \\
$\mathrm{AS}$                         & $11.186 \pm 2.050$  & $3.290$                   & $100$ & $-$ \\
$\mathrm{EAS}_{\mathrm{light}}$       & $9.781 \pm 4.906$   & $8.269 \times 10^{-1}$    & $100$ & $-$ \\
$\mathrm{EAS}$                        & $9.236 \pm 4.504$   & $3.251$                   & $100$ & $-$ \\
$\mathrm{PFACO}_{\mathrm{light}}$     & $8.934 \pm 4.145$   & $1.961 \times 10^{-1}$    & $100$ & $-$ \\
$\mathrm{PFACO}$                      & $8.722 \pm 3.873$   & $6.407 \times 10^{-1}$    & $100$ & $\approx$ \\
$\mathrm{STAPNet}$                    & $8.829 \pm 4.755$   & $ 4.767 \times 10^{-3}$    & $83$  & $\times$ \\
$\mathrm{ImiPath}^{\mathrm{light}}_{\mathrm{PFACO}}$   & $8.660 \pm 4.104$   & $\mathbf{1.506 \times 10^{-1}}$    & $100$ & $-$ \\
$\mathrm{ImiPath}_{\mathrm{PFACO}}$                    & $\mathbf{8.320 \pm 3.829}$   & $8.160 \times 10^{-1}$    & $100$ &  \\
\midrule

\multicolumn{5}{@{}l}{\textit{Map Scale: }Medium} \\
\midrule
$\mathrm{AS}_{\mathrm{light}}$        & $19.870 \pm 11.581$ & $2.168$                   & $100$ & $-$ \\
$\mathrm{AS}$                         & $18.166 \pm 10.332$ & $8.791$                   & $100$ & $-$ \\
$\mathrm{EAS}_{\mathrm{light}}$       & $16.588 \pm 9.536$  & $2.248$                   & $100$ & $-$ \\
$\mathrm{EAS}$                        & $15.548 \pm 8.921$  & $8.773$                   & $100$ & $-$ \\
$\mathrm{PFACO}_{\mathrm{light}}$     & $14.225 \pm 7.378$  & $1.499$                   & $100$ & $-$ \\
$\mathrm{PFACO}$                      & $13.919 \pm 7.023$  & $4.681$                   & $100$ & $\approx$ \\
$\mathrm{STAPNet}$                    & $10.792 \pm 6.805$  & $ 6.440 \times 10^{-3}$    & $58$  & $\times$ \\
$\mathrm{ImiPath}^{\mathrm{light}}_{\mathrm{PFACO}}$   & $13.704 \pm 7.293$  & $\mathbf{7.409 \times 10^{-1}}$    & $100$ & $-$ \\
$\mathrm{ImiPath}_{\mathrm{PFACO}}$                    & $\mathbf{13.234 \pm 6.922}$  & $2.186$                   & $100$ &  \\
\bottomrule
\end{tabular*}

\vspace{1mm}
\noindent
\begin{minipage}{\tblwidth}
\footnotesize
\raggedright
* For stochastic baselines, the subscript ``light'' denotes the lightweight configuration
(population size = 15, search iterations = 10), while the notation without a subscript
denotes the full configuration (population size = 30, search iterations = 20).\\
* ``$+$'': baseline better; ``$-$'': ImiPath better; ``$\approx$'': no significant difference;
``$\times$'': test not conducted due to insufficient successful trials.
\end{minipage}
\end{table}

\begin{table}[width=\linewidth,cols=5,pos=t]
\caption{Comparison of STAPNet, ImiPath, and representative stochastic path planners on large map scales under local FoV constraints.}
\label{tab:aco_large}

\begin{tabular*}{\tblwidth}{@{} LCCCC @{}}
\toprule
\textit{Method} & \textit{APL $\pm$ SD} & \textit{Time(s)} & \textit{SR(\%)} & \textit{$\mathrm{APL}_{\mathrm{p}}$} \\
\midrule
\multicolumn{5}{@{}l}{$\textit{Map Scale: }Medium-Large$} \\
\midrule
$\mathrm{AS}_{\mathrm{light}}$        & $24.996 \pm 12.909$ & $2.536$                   & $100$ & $-$ \\
$\mathrm{AS}$                         & $22.917 \pm 12.197$ & $9.848$                   & $100$ & $-$ \\
$\mathrm{EAS}_{\mathrm{light}}$       & $20.273 \pm 10.988$ & $2.493$                   & $100$ & $-$ \\
$\mathrm{EAS}$                        & $18.529 \pm 9.942$  & $9.869$                   & $100$ & $-$ \\
$\mathrm{PFACO}_{\mathrm{light}}$     & $16.492 \pm 8.314$  & $4.539 \times 10^{-1}$    & $100$ & $-$ \\
$\mathrm{PFACO}$                      & $16.062 \pm 8.210$  & $1.410$                   & $100$ & $\approx$ \\
$\mathrm{STAPNet}$                    & $14.790 \pm 8.458$  & $ 7.843 \times 10^{-3} $    & $56$  & $\times$ \\
$\mathrm{ImiPath}^{\mathrm{light}}_{\mathrm{PFACO}}$   & $15.782 \pm 8.134$  & $\mathbf{5.355 \times 10^{-1}}$    & $100$ & $-$ \\
$\mathrm{ImiPath}_{\mathrm{PFACO}}$                    & $\mathbf{15.448 \pm 7.950}$  & $1.426$                   & $100$ &  \\
\midrule

\multicolumn{5}{@{}l}{$\textit{Map Scale: }Large$} \\
\midrule
$\mathrm{AS}_{\mathrm{light}}$        & $31.264 \pm 17.812$ & $3.960$                   & $100$ & $-$ \\
$\mathrm{AS}$                         & $29.045 \pm 16.376$ & $2.905 \times 10^{1}$     & $100$ & $-$ \\
$\mathrm{EAS}_{\mathrm{light}}$       & $25.824 \pm 15.221$ & $3.858$                   & $100$ & $-$ \\
$\mathrm{EAS}$                        & $23.647 \pm 13.483$ & $15.334$                  & $100$ & $-$ \\
$\mathrm{PFACO}_{\mathrm{light}}$     & $20.383 \pm 11.121$ & $6.376 \times 10^{-1}$    & $100$ & $-$ \\
$\mathrm{PFACO}$                      & $20.142 \pm 10.304$ & $1.815$                   & $100$ & $\approx$ \\
$\mathrm{STAPNet}$                    & $18.694 \pm 13.731$ & $ 9.401 \times 10^{-3} $    & $54$  & $\times$ \\
$\mathrm{ImiPath}^{\mathrm{light}}_{\mathrm{PFACO}}$   & $19.577 \pm 10.973$ & $\mathbf{7.617 \times 10^{-1}}$    & $100$ & $-$ \\
$\mathrm{ImiPath}_{\mathrm{PFACO}}$                    & $\mathbf{19.097 \pm 10.330}$ & $1.895$                   & $100$ &  \\
\bottomrule
\end{tabular*}

\vspace{1mm}
\noindent
\begin{minipage}{\tblwidth}
\footnotesize
\raggedright
* ``$+$'': baseline better; ``$-$'': ImiPath better; ``$\approx$'': no significant difference;
``$\times$'': test not conducted due to insufficient successful trials.
\end{minipage}
\end{table}

\begin{table}[width=\linewidth,cols=6,pos=t]
\caption{Comparison of ImiPath with deterministic planners across different map scales.}
\label{tab:astar}
\centering
\footnotesize
\setlength{\tabcolsep}{3pt}
\renewcommand{\arraystretch}{0.95}
\resizebox{0.98\linewidth}{!}{%
\begin{tabular}{@{} lccccc @{}}
\toprule
\textit{Method} & \textit{APL $\pm$ SD} & \textit{Time(s)} & \textit{SR(\%)} & \textit{Explored Nodes} & \textit{$\mathrm{APL}_{\mathrm{p}}$} \\
\midrule

\multicolumn{6}{@{}l}{\textit{Map Scale: FoV Scale}} \\
\midrule
$\mathrm{A}^{*}_{\mathrm{Global}}$ & $4.591 \pm 2.621$ & $1.064 \times 10^{-4}$ & $100$ & $6.67 \pm 5.099$ & $\approx$ \\
$\mathrm{WA}^{*}\mathrm{+CF}$      & $5.021 \pm 3.228$ & $1.018 \times 10^{-2}$ & $100$ & $6.37 \pm 5.094$ & $\approx$ \\
$\mathrm{FS+PPM}$                  & $4.957 \pm 3.131$ & $1.154 \times 10^{-2}$ & $100$ & $13.03 \pm 13.782$ & $\approx$ \\
$\mathrm{STAPNet}$                 & $4.742 \pm 2.943$ & $1.133 \times 10^{-2}$ & $95$  & $6.63 \pm 5.745$ & $\times$ \\
$\mathrm{ImiPath}_{ \mathrm{A}^{*}_{\mathrm{Global}} }$ & $4.868 \pm 3.268$ & $1.090 \times 10^{-2}$ & $100$ & $6.70 \pm 5.652$ &  \\
\midrule

\multicolumn{6}{@{}l}{\textit{Map Scale: Small}} \\
\midrule
$\mathrm{A}^{*}_{\mathrm{Local}}$  & $9.318 \pm 4.178$ & $2.285 \times 10^{-3}$ & $100$ & $40.68 \pm 31.308$ & $-$ \\
$\mathrm{ImiPath}_{\mathrm{A}^{*}_{\mathrm{Local}}}$ & $\mathbf{8.361 \pm 3.717}$ & $7.430 \times 10^{-2}$ & $100$ & $\mathbf{6.29 \pm 13.110}$ &  \\
\midrule

\multicolumn{6}{@{}l}{\textit{Map Scale: Medium}} \\
\midrule
$\mathrm{A}^{*}_{\mathrm{Local}}$  & $15.266 \pm 6.933$ & $6.842 \times 10^{-3}$ & $100$ & $98.30 \pm 69.814$ & $\approx$ \\
$\mathrm{ImiPath}_{\mathrm{A}^{*}_{\mathrm{Local}}}$ & $\mathbf{13.307 \pm 6.804}$ & $1.936 \times 10^{-1}$ & $100$ & $\mathbf{71.93 \pm 81.371}$ &  \\
\midrule

\multicolumn{6}{@{}l}{\textit{Map Scale: Medium-large}} \\
\midrule
$\mathrm{A}^{*}_{\mathrm{Local}}$  & $16.861 \pm 6.834$ & $1.078 \times 10^{-2}$ & $100$ & $184.584 \pm 115.388$ & $-$ \\
$\mathrm{ImiPath}_{\mathrm{A}^{*}_{\mathrm{Local}}}$ & $\mathbf{14.628 \pm 6.683}$ & $3.025 \times 10^{-1}$ & $100$ & $\mathbf{54.86 \pm 55.953}$ &  \\
\midrule

\multicolumn{6}{@{}l}{\textit{Map Scale: Large}} \\
\midrule
$\mathrm{A}^{*}_{\mathrm{Local}}$  & $20.884 \pm 8.735$ & $1.839 \times 10^{-2}$ & $100$ & $242.61 \pm 184.584$ & $-$ \\
$\mathrm{ImiPath}_{\mathrm{A}^{*}_{\mathrm{Local}}}$ & $\mathbf{17.629 \pm 8.557}$ & $7.300 \times 10^{-1}$ & $100$ & $\mathbf{90.05 \pm 93.796}$ &  \\
\bottomrule
\end{tabular}
}
\end{table}

\begin{figure} 
  \centering
  \includegraphics[width=0.9\linewidth]{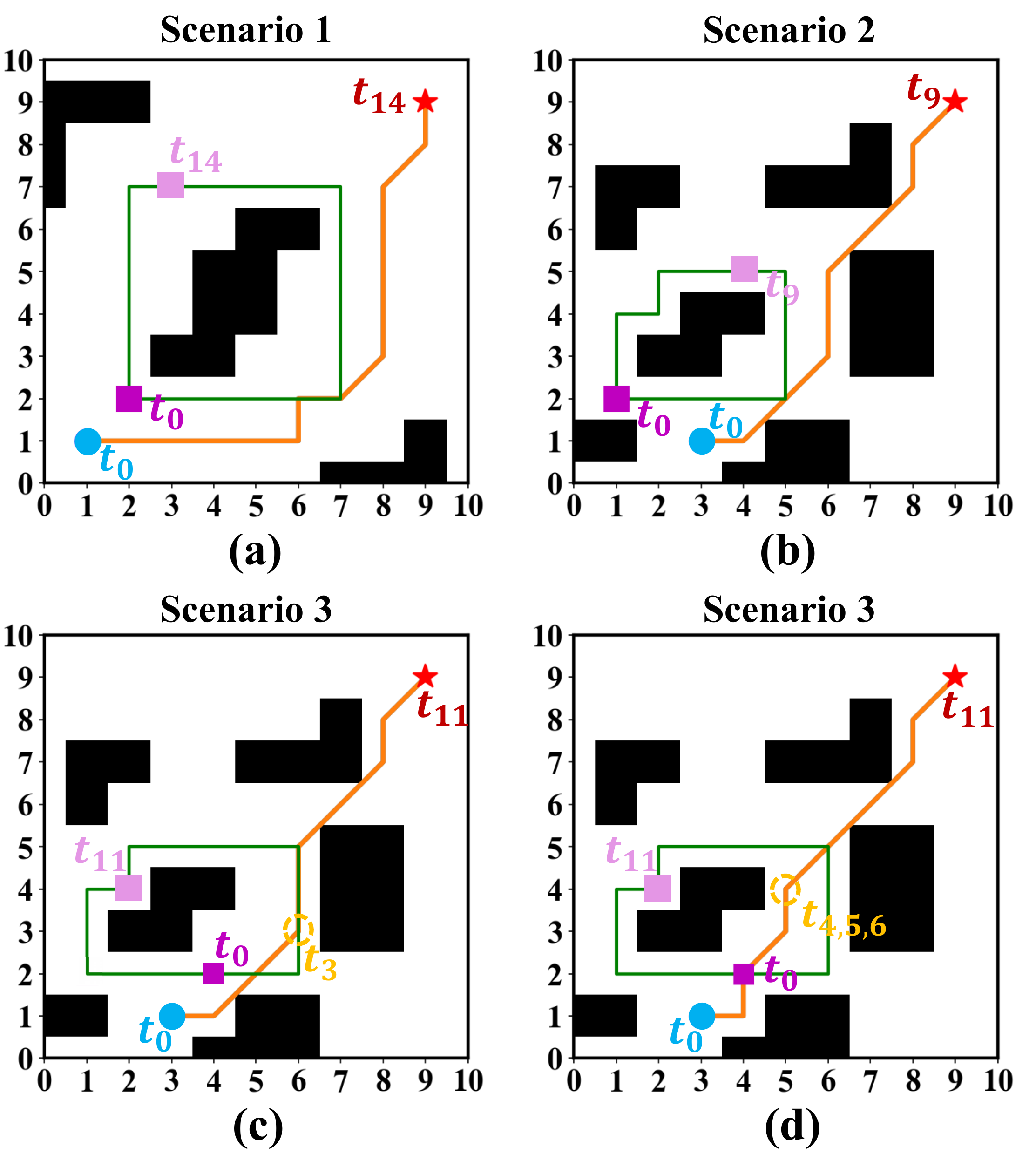}
  \caption{
 Dynamic-environment results on fixed-size FoV maps. The blue dot and red star denote the start and goal, respectively, and the orange polyline denotes the robot trajectory. The dark-purple and light-purple squares indicate the obstacle positions at \(t_{0}\) and \(t_{N}\), respectively, while the green polyline shows the obstacle trajectory. Yellow circles mark key events. (a) and (b) show the same task under different congestion levels. (c) illustrates a collision-risk case, and (d) shows the corresponding conflict-resolution result generated by ImiPath.
  }
  \label{fig:dynamic}
\end{figure}

\subsection{Simulation Results}

\subsubsection{Comparisons With Baselines Across Different Map Scales}

This subsection compares ImiPath with representative stochastic and deterministic planners under $11\times11$ FoV across map scales. 
STAPNet is additionally included as a standalone policy baseline without explicit search, whereas ImiPath denotes the planner-guided variants using STAPNet-derived priors. 
All methods were evaluated on the same set of 100 randomly sampled start--goal instances from held-out test maps. 
These test instances were disjoint from the expert demonstrations used for training, and no test map was involved in model selection or hyperparameter tuning.

Tables~\ref{tab:aco_small_medium} and~\ref{tab:aco_large} summarize the results for stochastic planners. 
$\mathrm{ImiPath}_{\mathrm{PFACO}}$ maintains a 100\% success rate across all scales and achieves the best or statistically comparable APL among the stochastic methods. 
These results indicate that integrating the prior into stochastic search improves robustness and preserves path quality when transferring to larger unseen maps under the same local-FoV constraint.
Moreover, $\mathrm{ImiPath}^{\mathrm{light}}_{\mathrm{PFACO}}$ achieves competitive or superior path quality and computation time compared with several full-configuration baselines, even under a lightweight configuration. This demonstrates that the spatiotemporal priors provide effective directional guidance for search, enabling ImiPath to reduce redundant exploration and achieve a favorable quality--efficiency tradeoff.

Table~\ref{tab:astar} reports the deterministic comparisons. 
On maps larger than the local FoV, the advantage of the $A^*_{\mathrm{Local}}$-based $\mathrm{ImiPath}_{\mathrm{A}^{*}_{\mathrm{Local}}}$ variant becomes more evident. Compared with $A^*_{\mathrm{Local}}$, $\mathrm{ImiPath}_{\mathrm{A}^{*}_{\mathrm{Local}}}$ consistently maintains a 100\% success rate, achieves shorter or comparable paths, and explores substantially fewer nodes on most map scales. 
This suggests that the prior offers more informative local search guidance than goal-direction cues alone, enabling the planner to suppress redundant exploration and mitigate myopic decisions under partial observability. 
Although $\mathrm{ImiPath}_{\mathrm{A}^{*}_{\mathrm{Local}}}$ introduces additional computation due to neural inference and prior-guided search, its runtime remains practical for online planning, ranging from $1.090 \times 10^{-2}$ s per planning instance on the FoV-scale benchmark to $7.300 \times 10^{-1}$ s on large maps.
Overall, these results demonstrate that ImiPath improves upon the local deterministic baseline, with particularly clear benefits in larger-scale local path planning scenarios.

On FoV scale, $\mathrm{ImiPath}_{\mathrm{A}^{*}_{\mathrm{Local}}}$ achieves a 100\% success rate, competitive or slightly better path quality than the compared methods, and explored-node counts close to those of $A^*_{\mathrm{Global}}$ and WA*+CF. 
On the FoV scale, $\mathrm{ImiPath}_{\mathrm{A}^{*}_{\mathrm{Local}}}$ achieves a 100\% success rate, competitive or slightly better path quality than the compared methods, and explored-node counts close to those of $A^*_{\mathrm{Global}}$ and WA*+CF. 
These results suggest that $\mathrm{ImiPath}_{\mathrm{A}^{*}_{\mathrm{Local}}}$ can approximate expert-derived search preferences and maintain a compact search process even under local observations. 
More importantly, its advantage becomes more pronounced on maps larger than the local FoV, where the learned spatiotemporal prior provides more informative directional guidance than short-range goal cues alone, thereby reducing redundant exploration in local path planning while preserving path quality and success rate.

\subsubsection{ImiPath in Dynamic Environments}

This subsection evaluates the adaptability of ImiPath in dynamic environments. 
As shown in Fig.~\ref{fig:dynamic}(a) and (b), ImiPath successfully completes the same navigation task under different congestion levels while maintaining collision-free trajectories with respect to the moving obstacle. 
Fig.~\ref{fig:dynamic}(c) illustrates a more challenging case in which the trajectory planned under a static assumption would intersect the obstacle path. 
Because ImiPath predicts actions from the current local observation, it can update the planning process online in response to environmental changes. 
For potential conflict events, the action space is extended with a stop action, allowing the robot to wait temporarily when the next movement may lead to collision and to resume motion after the obstacle has passed. 
As shown in Fig.~\ref{fig:dynamic}(d), this mechanism enables ImiPath to resolve the conflict online and continue toward the goal.
The corresponding experimental videos are provided in the supplementary materials.

The quantitative results in Table~\ref{tab:dynamic} further demonstrate the reliability of ImiPath in dynamic scenarios. 
Across all three scenarios, the method achieves a 100\% success rate, with average path lengths ranging from 11.071 to 14.828 and planning times between $5.147\times10^{-2}$ s and $6.740\times10^{-2}$ s. 
The number of explored nodes remains no greater than 15 in all cases, indicating efficient online search under local observability. 
Overall, these results show that ImiPath can make situation-aware decisions and generate reliable paths in dynamic environments through detouring and temporary waiting when necessary.

\begin{table}[width=\linewidth,cols=5,pos=t]
\caption{Planning results of ImiPath in multiple dynamic scenarios.}
\label{tab:dynamic}
\begin{tabular*}{\tblwidth}{@{} CCCCC @{}}
\toprule
\textit{Scenario} & \textit{APL} & \textit{Time(s)} & \textit{SR(\%)} & \textit{Explored Nodes} \\
\midrule
Scenario 1 & 14.828 & $5.147 \times 10^{-2}$ & 100 & 15 \\
Scenario 2 & 11.071 & $5.934 \times 10^{-2}$ & 100 & 10 \\
Scenario 3 & 13.071 & $6.740 \times 10^{-2}$ & 100 & 10 \\
\bottomrule
\end{tabular*}
\end{table}

\subsubsection{Ablation studies}\label{Ablation}




\noindent\textbf{Effect of Training Data Scale on Prior Learning}
To evaluate the effect of dataset scale on prior learning, five training sets containing $1 \times 10^{4}$, $2 \times 10^{4}$, $1 \times 10^{5}$, $2 \times 10^{5}$, and $3 \times 10^{5}$ demonstrations were constructed using the same expert planner. 

\begin{figure} 
\centering
\includegraphics[width=0.9\columnwidth]{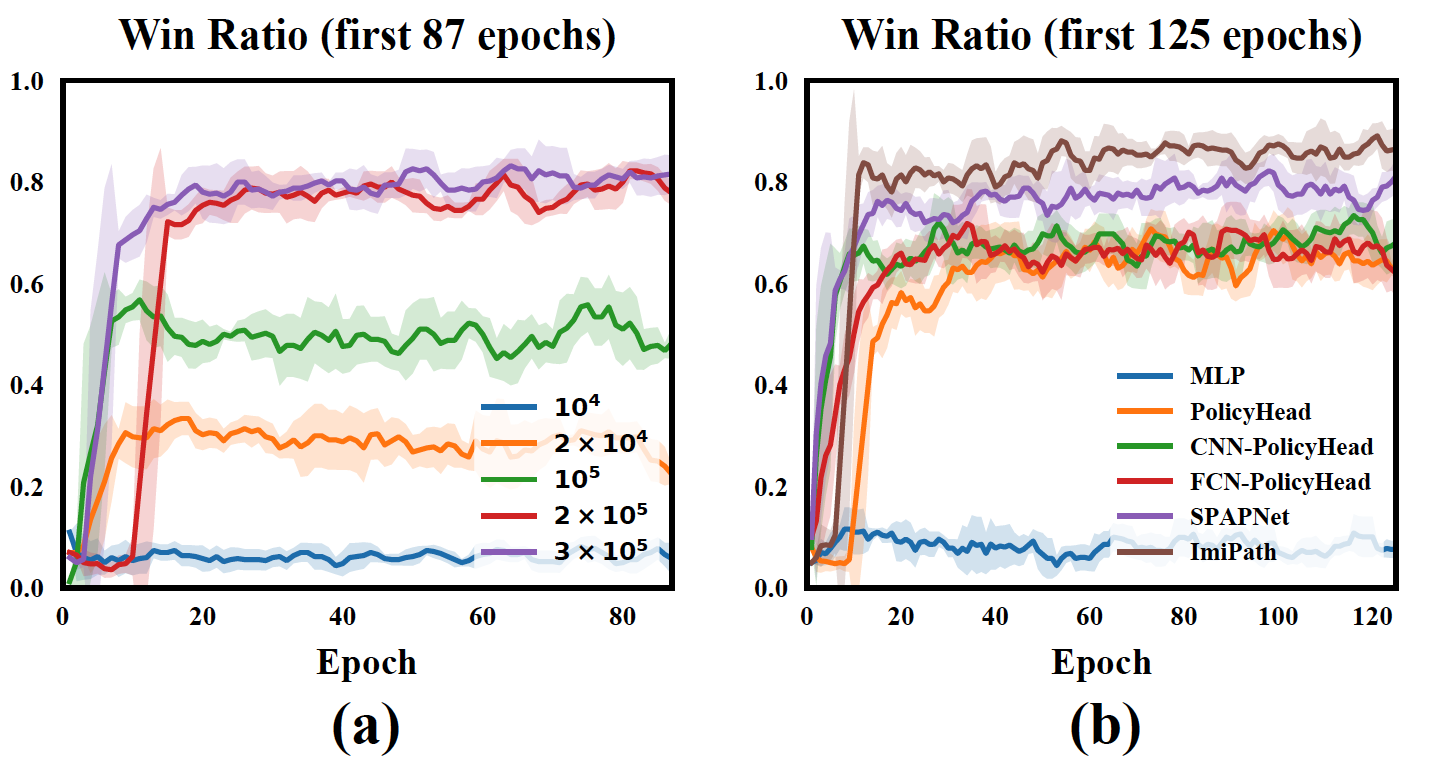}
\caption{
Effects of dataset scale and network architecture on decision-making performance.
(a) Training performance of STAPNet under different dataset sizes.
(b) Performance comparison of different network architectures and the complete ImiPath framework.
Shaded areas indicate the 95\% confidence intervals.
}
\label{fig:ablation23}
\end{figure}

In this experiment, the win ratio (WR) is used as the evaluation metric to measure the decision quality of the learned policy. 
Specifically, a test instance is counted as a win if the action sequence generated under the learned policy achieves better planning performance than the reference baseline according to the predefined evaluation criterion. 
The win ratio is defined as $WR=\frac{N_{\mathrm{win}}}{N_{\mathrm{total}}}$, where $N_{\mathrm{win}}$ denotes the number of winning test instances and $N_{\mathrm{total}}$ denotes the total number of test instances. 
In this experiment, $N_{\mathrm{total}}=100$.
Fig.~\ref{fig:ablation23}(a) shows the training curves of STAPNet under different dataset sizes. 
The vertical axis reports the win ratio, and the horizontal axis denotes the training epoch.

As shown in Fig.~\ref{fig:ablation23}(a), increasing the dataset size improves convergence stability and final decision quality, mainly because larger datasets provide more diverse state--action pairs for learning transferable local priors. 
When the dataset size reaches approximately $2 \times 10^{5}$ to $3 \times 10^{5}$, the performance curves begin to saturate and the differences among larger datasets become marginal, indicating diminishing returns from further increasing the number of demonstrations.


\noindent\textbf{Effect of Network Architecture on Prior Learning}
To assess the contribution of the STAPNet architecture, six variants were evaluated under identical training settings: MLP, PolicyHead only, CNN + PolicyHead, FCN + PolicyHead, STAPNet with the spatiotemporal encoder and policy head, and the complete ImiPath framework. 
Fig.~\ref{fig:ablation23}(b) compares their win ratios across training epochs.

The results show that STAPNet achieves a higher and more stable win ratio than the MLP, PolicyHead-only, CNN-based, and FCN-based variants. 
The PolicyHead-only model converges rapidly but saturates at a relatively low level, indicating that a prediction head without sufficient feature extraction capacity cannot effectively encode spatiotemporal planning cues. 
The CNN + PolicyHead and FCN + PolicyHead variants achieve intermediate performance, suggesting that spatial feature extraction alone is useful but insufficient for fully modeling the interaction between obstacle layout, goal direction, and motion history. 
The MLP baseline exhibits the weakest and least stable performance, further confirming the necessity of structured spatial-temporal representation learning.
ImiPath achieves the best overall decision performance, demonstrating that integrating STAPNet-derived priors into the planning pipeline further improves decision quality beyond standalone policy prediction. 
These results validate the effectiveness of the proposed spatiotemporal encoder and its role in learning transferable local decision priors.

\begin{table}[width=\linewidth,cols=5,pos=h]
\caption{Ablation results of different prior-fusion strategies for deterministic ImiPath.}
\label{tab:hybrid_type2}

\begin{tabular*}{\tblwidth}{@{} LCCCC @{}}
\toprule
\textit{Method} & \textit{APL $\pm$ SD} & \textit{Time(s)} & \textit{SR(\%)} & \textit{$\mathrm{APL}_{\mathrm{p}}$} \\
\midrule
$\mathrm{STAPNet}$ 
& $4.742 \pm 2.943$ 
& $1.133 \times 10^{-2}$ 
& $95$ 
& $\times$ \\

$\mathrm{ImiPath}_{\left(f(n)_1\right)}$ 
& $5.895 \pm 3.736$ 
& $1.172 \times 10^{-2}$ 
& $100$ 
& $-$ \\

$\mathrm{ImiPath}_{\left(f(n)_2\right)}$ 
& $5.501 \pm 3.369$ 
& $1.910 \times 10^{-2}$ 
& $100$ 
& $-$ \\

$\mathrm{ImiPath}_{\left(f(n)_3\right)}$ 
& $\mathbf{4.868 \pm 3.268}$ 
& $1.090 \times 10^{-2}$ 
& $100$ 
& \\
\bottomrule
\end{tabular*}
\end{table}

\begin{table}[width=\linewidth,cols=5,pos=h]
\caption{Ablation results of different prior-fusion strategies for stochastic ImiPath.}
\label{tab:hybrid_type1}

\begin{tabular*}{\tblwidth}{@{} LCCCC @{}}
\toprule
\textit{Method} & \textit{APL $\pm$ SD} & \textit{Time(s)} & \textit{SR(\%)} & \textit{$\mathrm{APL}_{\mathrm{p}}$} \\
\midrule

\multicolumn{5}{@{}l}{\textit{No Fusion}} \\
\midrule

$ \mathrm{STAPNet}$ 
& $5.127 \pm 3.951$ 
& $3.365 \times 10^{-1}$ 
& $99$ 
& $\times$ \\

\midrule
\multicolumn{5}{@{}l}{\textit{Hybrid v1}} \\
\midrule

$\mathrm{ImiPath}_{\mathrm{PFACO}}$ 
& $4.912 \pm 3.340$ 
& $9.841 \times 10^{-1}$ 
& $100$ 
& $-$ \\

\midrule
\multicolumn{5}{@{}l}{\textit{Hybrid v2}} \\
\midrule

$\mathrm{ImiPath}_{\mathrm{PFACO}}$ 
& $4.767 \pm 1.580$ 
& $3.344 \times 10^{-1}$ 
& $100$ 
& \\

\bottomrule
\end{tabular*}
\end{table}

\noindent\textbf{Effect of Prior-Guided Planning Strategies}\label{fig:ablationAco}
To further evaluate the effectiveness of the proposed prior-guided planning mechanism, ablation experiments were conducted for both deterministic and stochastic variants of ImiPath on Fov scale maps. 
Tables~\ref{tab:hybrid_type2} and~\ref{tab:hybrid_type1} summarize the results of different prior-fusion strategies.
 
For the deterministic A*-based variant, STAPNet predictions were incorporated into the node evaluation function. 
Three formulations were compared:
$f(n)_1=g(n)+h_{\mathrm{local}}(n)+\phi$, 
$f(n)_2=g(n)+\rho h_{\mathrm{local}}(n)+\mu(1-\phi)$, and 
$f(n)_3=g(n)+\rho h_{\mathrm{local}}(n)-\mu \log(\phi+\epsilon)$, 
where $\rho=\mu=0.5$. 
As shown in Table~\ref{tab:hybrid_type2}, $\mathrm{ImiPath}_{(f(n)_3)}$ achieves the shortest average path length while maintaining a 100\% success rate. 
This suggests that the logarithmic policy-prior term provides a more discriminative directional bias than the linear prior penalty. 
Specifically, $-\log(\phi+\epsilon)$ strongly penalizes actions assigned with low prior probabilities and gradually saturates for high-probability actions, which helps suppress unreliable node expansions while preserving flexibility among expert-preferred directions.
Accordingly, $\mathrm{ImiPath}_{(f(n)_3)}$ is adopted as the deterministic prior-fusion strategy in the main experiments.

For the stochastic ACO-based variant, two hybrid strategies were evaluated. 
Hybrid v1 initializes the pheromone matrix using STAPNet predictions before search, whereas Hybrid v2 incorporates the predicted policy into the transition probability at each iteration. 
As shown in Table~\ref{tab:hybrid_type1}, Hybrid v2 achieves the best overall performance, particularly for the PFACO-based variant, where it yields the shortest path length and a 100\% success rate. 
This indicates that iterative policy fusion is more effective than one-time prior initialization because the prior can continuously guide action selection during stochastic search.


\subsection{Experimental Results on the Magnetic Microrobot Platform}\label{sec:microrobot}

\begin{figure} 
\centering
\includegraphics[width=0.9\linewidth]{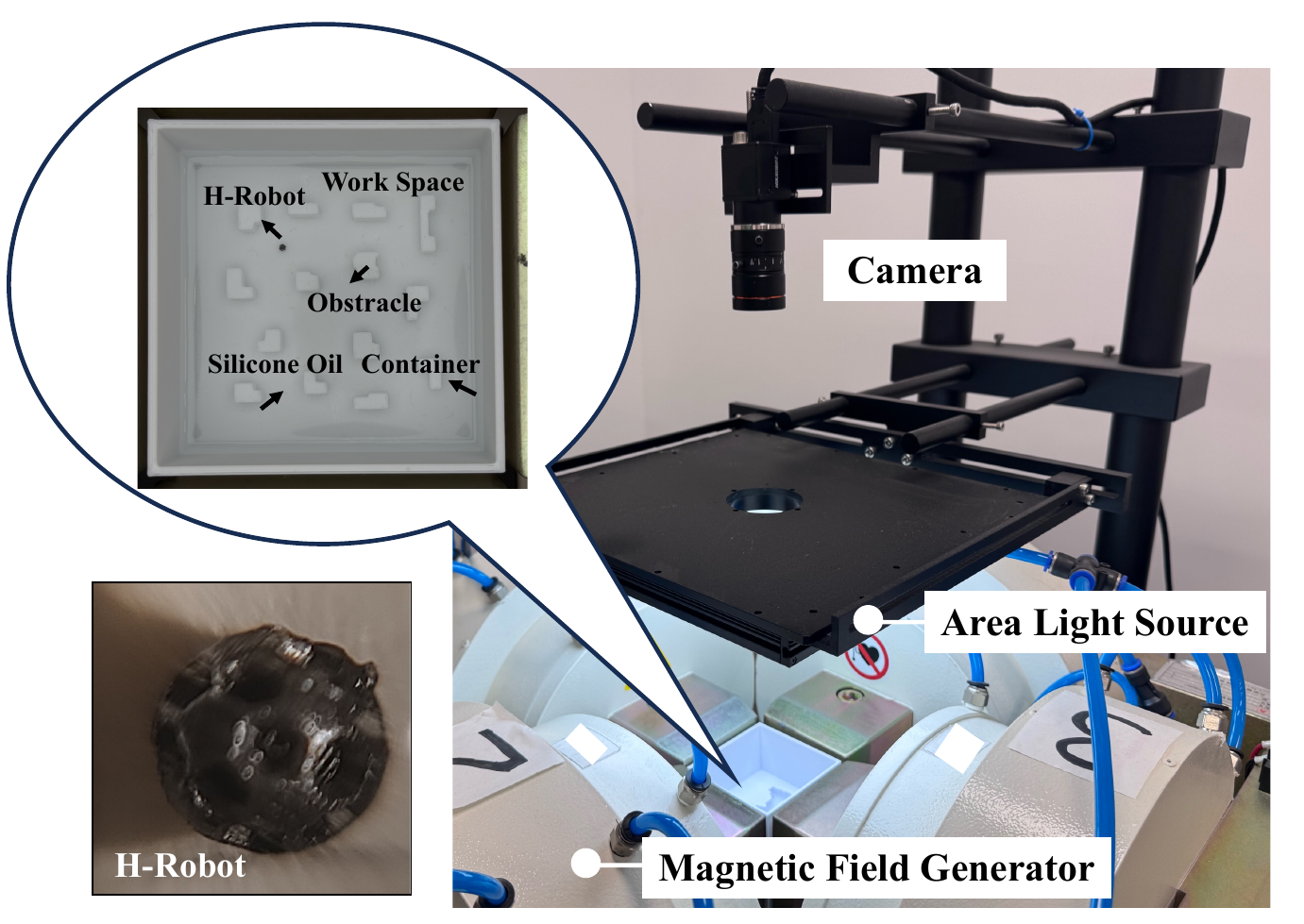}
\caption{ Experimental setup of the vision-based magnetic microrobot navigation system.
The platform integrates an industrial camera for real-time visual tracking, an area light source for uniform illumination, and a magnetic field generator for precise actuation. 
The inset illustrates the silicone oil container with obstacle configurations defining the navigation workspace, and the bottom-left image shows the fabricated H-robot. }
\label{fig:entity platform}
\end{figure}

\begin{figure} 
\centering
\includegraphics[width=0.85\linewidth]{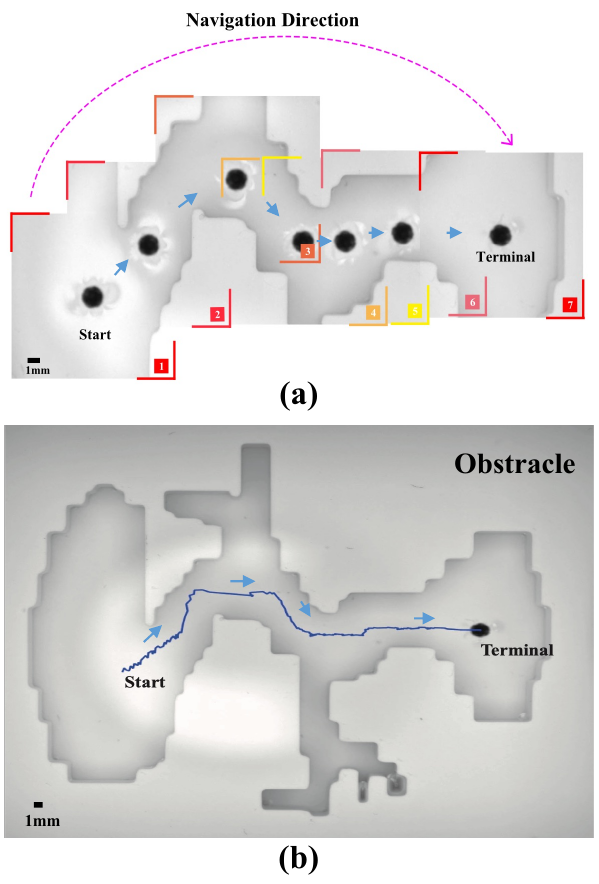}
\caption{Experimental results of ImiPath on the magnetic microrobot platform. (a) Composite visualization reconstructed from sequential local FoV observations collected at different intermediate positions along the navigation path; colored boxes denote the corresponding local observations. (b) Complete navigation trajectory generated by ImiPath from the start to the goal in a complex channel environment.}
\label{fig:entityexperiment}
\end{figure}

To assess the practical applicability of ImiPath under local observability, the proposed framework was further evaluated on a vision-based magnetic microrobot platform~\citep{zou2025modified}, as illustrated in Fig.~\ref{fig:entity platform}. 
The platform consists of three main components: a visual perception module, an illumination module, and a magnetic actuation module. 
An industrial camera mounted above the workspace provides real-time visual feedback for microrobot tracking, an area light source ensures stable image acquisition, and a magnetic field generator produces controlled magnetic fields for actuation.

The experiments were conducted in a silicone-oil container that simulates a low-Reynolds-number fluid environment. 
The container was fabricated by 3D printing with integrated obstacle structures, forming a constrained navigation workspace for the H-shaped microrobot shown in Fig.~\ref{fig:entity platform}. 
During execution, the robot position was continuously captured by the vision system and fed back to the controller for closed-loop navigation.
 
The experimental results are presented in Fig.~\ref{fig:entityexperiment}. 
Fig.~\ref{fig:entityexperiment}(a) shows a composite visualization reconstructed from sequential local-FoV observations collected along the navigation process, while Fig.~\ref{fig:entityexperiment}(b) shows the complete navigation trajectory generated by ImiPath from the start to the goal. 
These results demonstrate that ImiPath can support physically executable navigation using sequential local observations in a constrained channel environment. 
This results provide preliminary evidence of the practical deployability of the proposed framework in microrobotic navigation scenarios.

\section{Conclusion}\label{sec:conclusion}

This paper presented ImiPath, a prior-guided framework for path planning under partial observability. 
It distills reusable spatiotemporal decision priors from expert demonstrations and formulates them as local directional guidance for heterogeneous planning paradigms.
This design allows the priors to guide different planners, thereby reducing redundant search and improving planning efficiency.
Experiments demonstrate competitive path quality, and improved search efficiency over the baselines. 
Physical experiments on a magnetic microrobot platform provide the practical deployability in constrained navigation scenarios. 
Future work will explore more complex dynamic environments and extensions to multi-robot coordination and longer-horizon planning.

\bibliographystyle{cas-model2-names}
\bibliography{bib}





\end{document}